\documentclass[10pt,twocolumn,letterpaper]{article}

\usepackage{wacv}
\usepackage{times}
\usepackage{epsfig}
\usepackage{graphicx}
\usepackage{amsmath}
\usepackage{amssymb}
\usepackage{soul}
\usepackage{multirow}
\usepackage{subcaption}

\usepackage{bm}
\graphicspath{ {images/} }

\usepackage{listings} 

\newcommand{\RNum}[1]{\uppercase\expandafter{\romannumeral #1\relax}}
\newcommand{\quotes}[1]{``#1''}

\def\wacvPaperID{014} 

\wacvfinalcopy 

\ifwacvfinal
\fi

\ifwacvfinal
\usepackage[breaklinks=true,bookmarks=false]{hyperref}
\else
\usepackage[pagebackref=true,breaklinks=true,colorlinks,bookmarks=false]{hyperref}
\fi

\ifwacvfinal
\else
\fi

\begin{document}

\title{Morph Detection Enhanced by Structured Group Sparsity}

\author{{Poorya Aghdaie, Baaria Chaudhary, Sobhan Soleymani, Jeremy Dawson, Nasser M. Nasrabadi}\\
West Virginia University

}

\maketitle

\begin{abstract}
   In this paper, we consider the challenge of face morphing attacks, which substantially undermine the integrity of face recognition systems such as those adopted for use in border protection agencies. Morph detection can be formulated as extracting fine-grained representations, where local discriminative features are harnessed for learning a hypothesis. To acquire discriminative features at different granularity as well as a decoupled spectral information, we leverage wavelet domain analysis to gain insight into the spatial-frequency content of a morphed face. As such, instead of using images in the RGB domain, we decompose every image into its wavelet sub-bands using 2D wavelet decomposition and a deep supervised feature selection scheme is employed to find the most discriminative wavelet sub-bands of input images. To this end, we train a Deep Neural Network (DNN) morph detector using the decomposed wavelet sub-bands of the morphed and bona fide images. In the training phase, our structured group sparsity-constrained DNN picks the most discriminative wavelet sub-bands out of all the sub-bands, with which we retrain our DNN, resulting in a precise detection of morphed images when inference is achieved on a probe image. The efficacy of our deep morph detector which is enhanced by structured group lasso is validated through experiments on three facial morph image databases, i.e., VISAPP17, LMA, and MorGAN. 
\end{abstract}

\section{INTRODUCTION}


Face forgery detection has gained momentum recently in the biometric community owing to its vast application, especially in commercial face recognition systems \cite{qian2020thinking, li2020face, zhao2021multi, seibold2017detection, li2020advancing, dang2020detection}. Photo-realistic forged images tamper with the functionality and integrity of security checkpoints, where, ideally, there must be a zero-tolerance policy to false acceptance \cite{neubert2019face, zhang2018face, ferrara2014magic, singh2019robust}. Introduced in \cite{ferrara2014magic}, facial morph images, as one of the categories of the forged face images, can bypass established automated face recognition systems, as well as border control officers, where both struggle to distinguish a bona fide image from a morphed one \cite{nightingale2021perceptual} due to delicacy in synthesizing morphed samples. Face morphing attacks are synthesized using two look-alike genuine images, for example one representing a criminal and one for an innocent subject, in which the final morphed image can be verified against both subjects. Two underlying genuine images are sampled from two distributions, and the resulting morphed sample is characterized using the blended features of the two genuine images. If we assume the support of high-dimensional genuine and morphed samples are two underlying embedding low-dimensional manifolds, morphed samples are intentionally crafted near the discriminating boundary of the two manifolds, which justifies its verifiability against both real subjects.

\begin{figure}[t]
\begin{center}
\includegraphics[width=.99\linewidth]{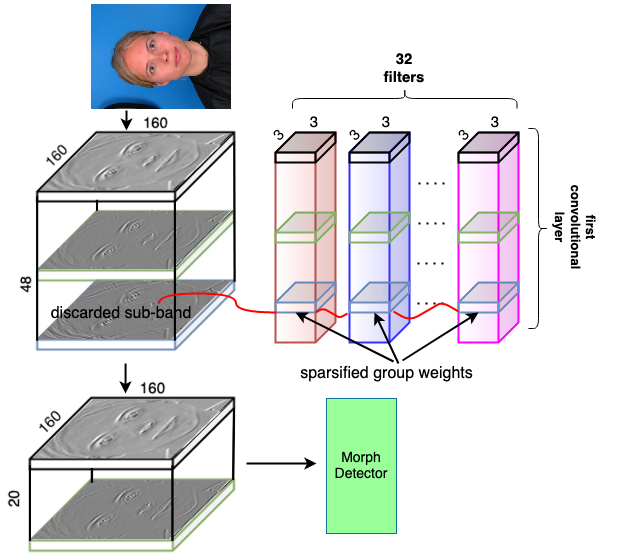}
\end{center}
   \caption{Group Lasso regularization, as a representation learning, leads to selecting the most discriminative sub-bands for detecting a morphed image.  }
   \label{Figure_1}
\end{figure}

Face morphing attacks are forged in either the image domain or in the latent domain. In image domain morphing, two genuine images are translated into a set of aligned averaged landmarks and a morphed sample is synthesized after warping and alpha blending. In latent space morphing, a generative adversarial network, which has an attached encoder, first captures the distribution of genuine images, and converts two genuine images into two latent vectors that can be mixed using the convex combination of both vectors. From the detection standpoint, previous research efforts have considered two approaches to detect morphed samples. Single morphed image detection, which is the focus of this study, labels an image as either genuine (negative) or morphed (positive). On the other hand, differential morph detection employs a probe image and an auxiliary one, which is usually a live photo of a subject, to define morph detection framework.

Deep learning-based techniques have shown compelling results for detecting morphed images through harnessing representation learning \cite{bengio2013representation} by mapping data samples into an embedding space where the separability between genuine and morphed samples is guaranteed through the aligning parameter space of a Deep Neural Network (DNN). Since the discrepancies between a bona fide and its corresponding morphed image are subtle and local, fine-grained feature learning \cite{hu2019see, zhao2021multi} can be tailored for our morph detection algorithm. To mitigate the curse of dimensionality, feature selection is a powerful tool to find the most discriminating hyperplane in a binary classification setting. A DNN can pick the most discriminative structured group features of its input space by adjusting its kernel parameters used in its first convolutional layer. Enforcing a group sparsity constraint over the weights of the first convolutional layer in a DNN through the manipulation of its loss function can guide the parameter space to convergence on a set of parameters that picks the most discriminative channels of input data (see Figure 1).

In this work, we tackle single image morph detection. Inspired by the aforementioned feature selection scheme, we investigate the application of group-L1 sparsity \cite{wen2016learning} over the weights of the first convolutional layer in a DNN as the criterion for selecting the most discriminative input samples' wavelet sub-bands. Discriminative wavelet sub-bands, which can be thought of as fine-grained features, are learned during the training of our DNN, guaranteeing that an optimal hyperplane will be found between genuine and morphed samples. We conduct experiments on three morph image datasets, i.e., VISAPP17 \cite{makrushin2017automatic}, LMA \cite{damer2018morgan}, and MorGAN \cite{damer2018morgan}. Given supervised data samples, our DNN sparsifies the kernel parameters of the first convolutional layer on-the-fly while training, which drives our sparsity-guided DNN into detecting morphed samples.

\section{RELATED WORKS}

\subsection{Morph Generation}
The landmark based image morphing technique \cite{makrushin2017automatic, damer2018morgan, seibold2018accurate, ferrara2014magic} considers the geometry of two bona fide subjects for generating a morphed sample. Every pixel location in both genuine images is warped to preserve the correspondence in the resulting morphed sample, and a convex combination of the warped pixels cross-dissolve the warped pixels in the real images to synthesize that pixel location in the final morphed sample. On the other hand, Generative Adversarial Network
(GAN)-based image morphing techniques focus on the distribution of real images. A trained GAN is able to find the distribution where data samples are drawn from. The point here is that a convex combination is generated in the latent domain $\mathcal{Z}$. Since GANs do not map an image into a latent vector $\mathcal{Z}$, an encoder attached to the generator of a GAN can achieve this mapping. A trained GAN maps two genuine images into a latent domain for interpolation, and a decoder maps the interpolated vector into the image domain to realize morphed samples \cite{damer2018morgan, abdal2019image2stylegan}.
\subsection{Morph Detection}
Different methods have been proposed for morph detection \cite{scherhag2019face, aghdaie2021detection, scherhag2017vulnerability, spreeuwers2018towards, scherhag2018performance, ferrara2019face, autherith2020detecting, venkatesh2021face}, some of which are discussed here. Ferrara \textit{et al.} \cite{ferrara2017face} introduced face demorphing to reverse the morphing process to detect altered images. Some research considers hand-crafted descriptors to detect morphed samples \cite{debiasi2019detection, 7791169, scherhag2018morph, 8272742}, where the fusion of different features has proven to be compelling for morph detection \cite{scherhag2018morph, scherhag2020face}. Deep morph detectors utilize the power of convolutional neural networks to detect morphed samples accurately \cite{ferrara2019face, raja2017transferable, seibold2017detection, seibold2018accurate, scherhag2019face, lu2020face}. Several studies investigate denoising methods for the task of morph attack detection. In \cite{scherhag2019detection}, Photo Response Non-Uniformity (PRNU) is investigated as an indicator to distinguish a bona fide sample from a morphed one. It is shown in \cite{scherhag2019detection} that morphed samples exhibit a distinguishing spectral response compared to the response related to bona fide images. In \cite{venkatesh2020detecting}, the residual noise of an image is utilized to extract deep features for training a morph detector. In \cite{aghdaie2021attention}, to increase the accuracy of morph detection, a soft attention mechanism is adopted to steer the focus of a DNN into the most discriminative spatial regions in an image under investigation. In \cite{blasingame2021leveraging}, a GAN-based single image morph discriminator is trained, which leverages adversarial learning to detect single image morphing attacks.

\subsection{Selecting Structured Features}
Feature selection is one of the most useful tools for classification, and finding relevant features that have the maximum mutual information with the predictor's output is highly desirable. When features are arranged in a group setting, group Lasso~\cite{yuan2006model}, as one of the many feature selection methods, has shown impressive proficiency \cite{zhang2012automatic, xiang2013efficient, wen2016learning, yang2014novel, gui2016feature}. When the $L_1$-norm of some structured parameters, such as grouped weights in a convolutional layer of a DNN, are added to the objective function in an optimization framework, which is known as Lasso regularization~\cite{tibshirani1996regression}, sparsified grouped parameters leads to feature selection, which is known as structured group sparsity in a DNN \cite{wen2016learning}.

\begin{figure*}[t]
\begin{center}
\includegraphics[width=.99\linewidth]{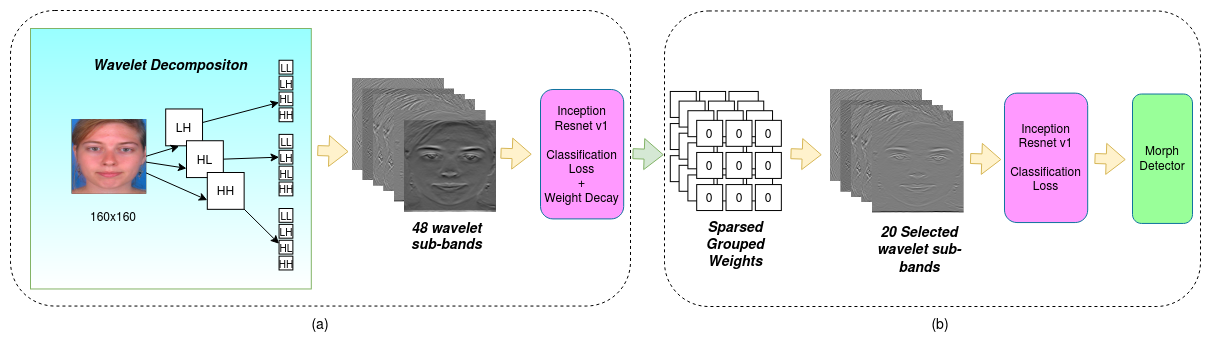}
\end{center}
  \caption{(a): Our modified Inception-ResNet-v1 \cite{szegedy2017inception} is trained with the 48-channel samples of both real and morphed images. Group Lasso constraint sparsifies the grouped weights of the first convolutional layer leading to sub-band selection.\\ 
  (b): We retrain our modified DNN using the 20 discriminative wavelet sub-bands, where the input channel size of our DNN is reduced to 20. A binary classifier is learned for morph detection.  }
\end{figure*}

\section{PROPOSED FRAMEWORK}

To incorporate fine-grained spatial-frequency features into our proposed framework, we leverage wavelet domain analysis as the basis of our deep morph detector. At varied granularity, wavelet sub-bands contain local discriminative information, where morphing artifacts are uncovered in this domain. Our wavelet-based deep morph detection mechanism is two-fold. First, we accomplish the sub-band selection to find the most informative subset of features to increase the confidence of our deep morph detector as far as inference is concerned. As presented in Figure 1, the optimization framework in the structured group sparsity zeros out the grouped weights corresponding to convolutional filters in a given layer of a DNN morph detector. Once some grouped weights in a convoluational layer converge to zero, those involved wavelet sub-bands are discarded, that is an implicit feature selection method. Secondly, we train our DNN detector using the selected wavelet sub-bands for the task of morph detection (see Figure 2).
\subsection{Sub-band Selection Based on Group Lasso}
We utilize the valuable spatial-frequency information provided by the wavelet decomposition to enhance the accuracy of our proposed morph detector. We pre-processed every input image using a three-level uniform wavelet decomposition. Since the morphing artifacts are revealed in the high frequency spectra, we discard the Low-Low (LL) sub-band after the first level of decomposition, and the remaining 48 wavelet sub-bands are extracted to be used by our deep morph detector. 

Since we intend to select the most discriminative wavelet sub-bands, our focus is on the first convolutional layer of our DNN. Please note that the input consists of $C$ wavelet sub-bands (channels) and the first  convolutional layer is defined in the space of $\mathcal{R}^{N \times C \times H \times V}$ where $N, C, H,$ and $V$ represent the number of filters, number of kernels, height, and width of a kernel, respectively. Here, we differentiate between the filter and the kernel terminologies as discussed in \cite{Bai:2019}. A kernel is a 2D array of size $H\times V$. A filter is an array of size $C\times H\times V$, which is a 3D structure of stacked 2D kernels. For every input image, we have 48 concatenated wavelet sub-bands as input to our DNN detector, which implies that the number of kernels in each filter of the first convolutional layer is equal to 48. We employ 32 different filters in the first convolutional layer, where the size of each kernel is $3\times3$. Thus, the dimensions of the first convolutional layer in the stage of feature selection are as follows: $N=32, C=48, H=3,$ and $V=3$. It is worth mentioning that we have 48 different grouped weights that are shown by $w_{l1} (:, c, :, :)$ for c $\in \{1,..., 48\}$, where the first layer weights are denoted by $w_{l1}$. 
Discriminative wavelet sub-bands are selected in accordance with a supervised feature selection algorithm. Our objective is to impose a group sparsity constraint on the parameters of the first convolutional layer in our DNN framework. Adding a structured sparsity regularization penalty to the classification loss of our DNN detector forces our network into sparsifying the grouped weights of the first convolutional layer, which implicitly results in discarding irrelevant wavelet sub-bands, or selecting the most discriminative sub-bands. Thus, only a certain number of informative wavelet sub-bands, out of 48, are selected, with which we train our DNN morph detector. 

\subsection{Rewriting the Classification Loss of the DNN Detector}
If we denote the classification loss of our DNN detector as $\mathcal{L}_{cl.}(w)$, the set of parameters of our network as $w$, the first layer weights as $w_{l1}$, and each grouped weight in the first layer as $w_{l1}^{(g)}$, the regularized loss function, denoted by $\mathcal{L_R}(w)$, for training the deep neural network is as follows:
\begin{equation}
\mathcal{L_R}(w) = \mathcal{L}_{cl.}(w)+\lambda \|w_{l1}\|_{1,2}= \mathcal{L}_{cl.}(w) +\lambda  \sum_{g \in \mathcal{G}_{l1} } \|w_{l1}^{(g)}\|_{2},
\end{equation}
where $\mathcal{G}_{l1}$ is a set composed of all the group weights of the convolutional filters in the first layer, and $\lambda$ is a parameter controlling the amount of sparsity. The regularized loss can be re-written as:
\begin{equation}
\mathcal{L_R}(w) = \mathcal{L}_{cl.}(w)+\lambda \sum_{c=1}^{C}\sqrt{\sum_{n=1}^{N}\sum_{h=1}^{H}\sum_{v=1}^{V} w_{l1}^2(n,c,h,v)},
\end{equation}
where $\mathcal{L}_{cl.}(w)$ delineates the binary cross-entropy classification loss and $N=32, C=48, H=3,$ and $V=3$.

\subsection{Learning Deep Morph Detector}
Regarding our feature selection scheme, as mentioned in Eqs. 1 and 2, there is a hyperparameter $\lambda$ in the optimization framework of our DNN detector, which is the regularization coefficient. To find the optimal regularization coefficient we utilize the validation sets of our datasets. To incorporate all three datasets in the selection process of hyperparameter $\lambda$, we combine all the images in the three datasets, i.e., VISAPP17 \cite{makrushin2017automatic}, LMA \cite{damer2018morgan}, and MorGAN \cite{damer2018morgan}, and we create a \quotes {universal dataset}. A hyperparameter search is performed over different values of $\lambda$. For each value of $\lambda$, our group sparsity-constrained DNN is trained using the training portion of the universal dataset, and the performance of the trained DNN detector is evaluated through Area Under the Curve (AUC) using the validation portion of the universal dataset. The highest AUC corresponds to the optimal value for $\lambda$. 
Once, we obtained the most discriminative wavelet sub-bands, we retrain our DNN using the selected wavelet sub-bands. Please note that since the number of input wavelet sub-bands is reduced due to feature selection, the number of channels $C$ in the filters of the first convolutional layer is also reduced.

\begin{figure}[t]
\begin{center}
\includegraphics[width=.65\linewidth]{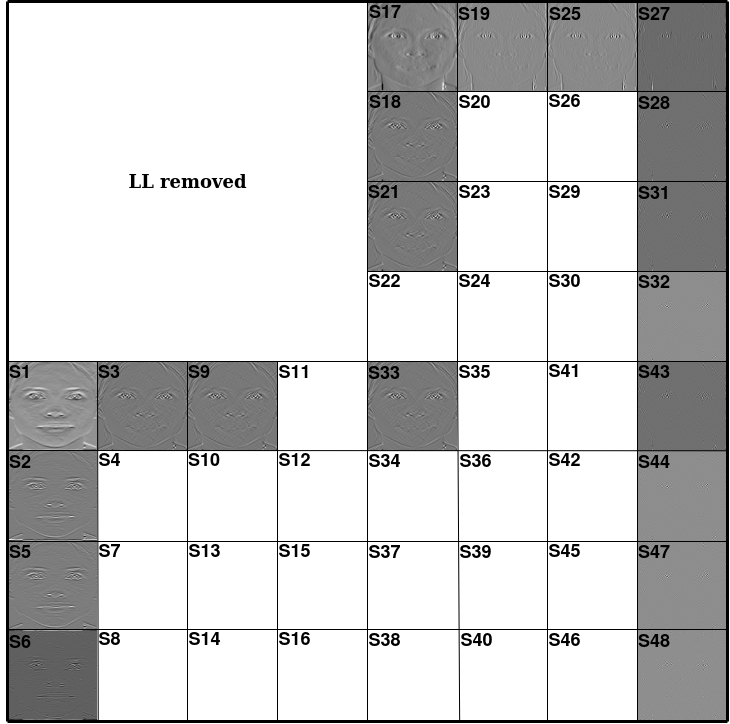} 
\end{center}
  \caption{Selected discriminative sub-bands using structured group sparsity. The white areas represent the irrelevant sub-bands that are discarded. The remaining informative sub-bands are displayed.}
\end{figure}

\begin{figure}[t]

    \includegraphics[width=0.99\linewidth]{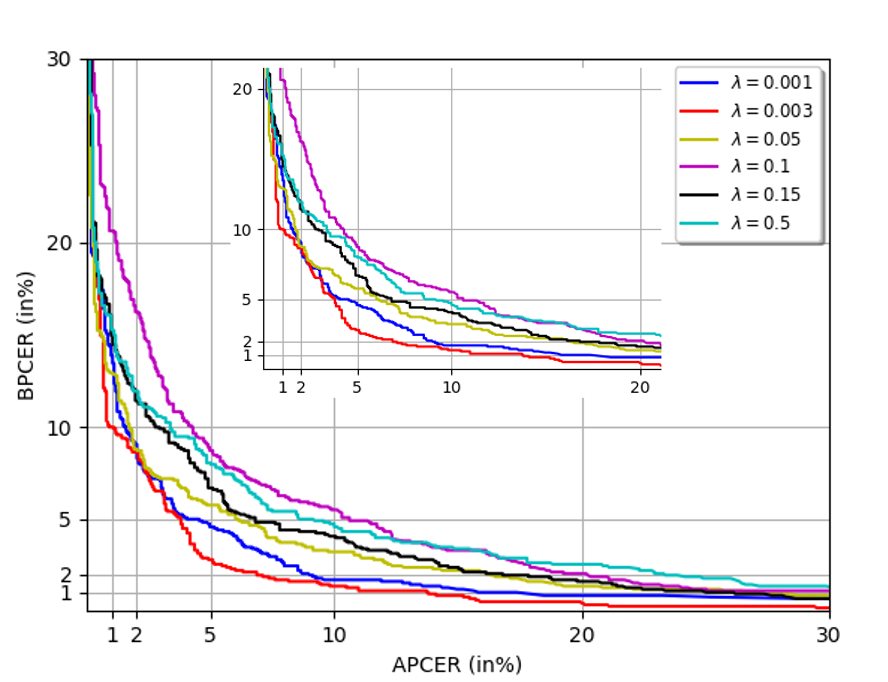} 
    \caption{DET curves corresponding to different values of $\lambda$  evaluated on the validation portion of the universal dataset.}
\end{figure}

\section{EVALUATIONS}
\subsection{Datasets}
We employ the VISAPP17 \cite{makrushin2017automatic}, MorGAN \cite{damer2018morgan}, and LMA \cite{damer2018morgan} datasets in this work. The VISAPP17 and LMA datasets were generated using facial landmark manipulation techniques, which is an alpha blending of the warped bona fide images. On the other hand, the MorGAN dataset is synthesized using a GAN, including a decoder network that utilizes transposed convolutional layers to transform a convex combination of two generated latent vectors into the image domain. The VISAPP17 dataset has been generated using the images in the Utrecht FCVP dataset, and the LMA and MorGAN datasets were generated using the CelebA dataset \cite{liu2015faceattributes}. For all images, face detection is performed via MTCNN and all images are resized to $160\times160$. We apply 2D undecimated wavelet decomposition on all images, and since we get 48 concatenated wavelet sub-bands for each image, the dimension of each data sample is $48\times160\times160$.

As for the universal dataset introduced in Section 3.4., the training set includes 1,631 bona fide, and 1,183 morphed images, the validation set consists of 462 bona fide, and 167 morphed samples and the test set includes 1,631 bonafide, and 1,183 morphed images.

\subsection{Experimental Setup}
We adopt a modified version of the Inception-ResNet-v1 \cite{szegedy2017inception} as the backbone of our DNN architecture for learning the discriminative sub-bands, as well as distinguishing morphed samples. We change the number of channels in the first convolutional layer of the Inception-ResNet-v1 to 48 during the sub-band selection stage. As mentioned in Section 3.2., a sample input consists of 48 stacked wavelet sub-bands; thus, the convolutional filters of the first layer have 48 channels, as seen in Figure 1. Inspired by \cite{wen2016learning}, we use group $L_{1}$-regularization to impose structured group sparsity constraint on the grouped weight parameters in the first convolutional layer of our deep neural network. For training our modified Inception-ResNet-v1, we adopt the Adam optimizer and training is done for 150 epochs. The learning schedule is as follows: the learning rate is initialized with 0.001, and it is divided by 10 after every 20 epochs. The training phase is accelerated using two 12 GB TITAN X (Pascal) GPUs.

\begin{figure}[t]
    \begin{center}
    \includegraphics[width=0.99\linewidth]{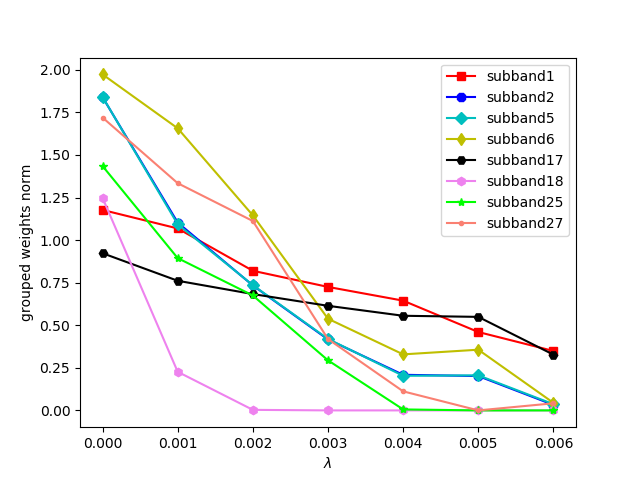} 
    \end{center}
   \caption{Displaying grouped weights decay related to some of the selected 20 sub-bands with respect to group sparsity hyperparameter $\lambda$.}

\end{figure}

\subsection{Evaluation Metrics}
We use the APCER, BPCER, and D-EER metrics to assess the performance of our deep morph detector. The first metric, Attack Presentation Classification Error Rate (APCER), is the percentage of morphed images that are classified as bona fide, and the second metric, Bona Fide Presentation Classification Error Rate (BPCER), represents percentage of bona fide samples that are classified as morphed. D-EER stands for Detection Equal Error Rate, at which APCER equals BPCER. The BPCER5 is the BPCER rate when APCER=5\%, and similarly the BPCER10 is the BPCER rate when APCER=10\%.

 \begin{table}[t]
\caption[Table caption text]{Performance of single morph detection: D-EER\%, BPCER@APCER=5\%, and BPCER@APCER=10\%.} 
\small
\begin{center}
\addtolength{\tabcolsep}{-0pt}
\begin{tabular}{ccccc}
\hline
Dataset&Algorithm&D-EER&5\%&10\%\\ 
\hline
\multirow{7}{*}{\rotatebox[origin=c]{90}{VISAPP17}}&BSIF+SVM~ \cite{kannala2012bsif}&                      16.51 &35.61&26.79\\
                                &SIFT+SVM~ \cite{lowe1999object}&     38.59 &82.40&75.60\\
                                &LBP+SVM~ \cite{ojala1994performance}&  38.00 &77.10 & 67.90\\
                                &SURF+SVM~ \cite{bay2006surf}&  30.45 &84.70 & 69.40\\
                                &RGB+DNN~ \cite{szegedy2017inception}&  1.76 &0.588 & 0.58\\
                                &48-sub-bands & 0.00 & 0.00 & 0.00\\
                                &\textbf{Ours}&                                 
                                {\bf0.00} & {\bf0.00} & {\bf0.00}\\

\hline

\multirow{7}{*}{\rotatebox[origin=c]{90}{LMA}}&BSIF+SVM~ \cite{kannala2012bsif}&                      33.05 &78.34&62.86\\
                                &SIFT+SVM~ \cite{lowe1999object}&     33.30 &83.40&72.00\\
                                &LBP+SVM~ \cite{ojala1994performance}&  28.00 &58.60 & 51.40\\
                                &SURF+SVM~ \cite{bay2006surf}&  37.40 &79.50 & 70.00\\
                                &RGB+DNN~ \cite{szegedy2017inception}&  9.10 &15.18 & 7.49\\
                                &48-sub-bands & 5.04 & 4.38 & 2.75\\
                                &\textbf{Ours}&                                 
                                {\bf 6.80} &{\bf 8.60}&{\bf 4.89}\\

\hline

\multirow{7}{*}{\rotatebox[origin=c]{90}{MorGAN}}&BSIF+SVM~\cite{kannala2012bsif}&                      1.57 &1.42&1.30\\
                                &SIFT+SVM~ \cite{lowe1999object}&     43.50 &93.20&84.20\\
                                &LBP+SVM~ \cite{ojala1994performance}&  20.10 &52.70 & 32.30\\
                                &SURF+SVM~ \cite{bay2006surf}&  39.95 &80.00 & 72.60\\
                                &RGB+DNN~ \cite{szegedy2017inception}&  2.44 &1.88 & 1.50\\
                                &48-sub-bands & 0.81 & 0.59 & 0.32\\
                                &\textbf{Ours}&                                 
                                {\bf0.42} &{\bf0.38}&{\bf0.22}\\

\hline

\end{tabular}
\end{center}

\label{table:results_cross}
\end{table}

\subsection{Tuning the Group Sparsity Regularization Hyperparameter $\lambda$ }
We performed a search over the group sparsity regularization coefficient $\lambda$. To find the optimal value for $\lambda$, as discussed in Section 3.4., we fine tune our modified Inception-ResNet-v1 DNN \cite{szegedy2017inception}, already pre-trained on VGGFace2, using the training portion of the universal dataset. For each value of hyperparameter $\lambda$, we trained our modified Inception-ResNet-v1 DNN using the training portion of the universal dataset, and we found the AUC metric when our trained DNN is evaluated using the validation portion of the universal dataset. It should be noted that we created the universal dataset for hyperparameter selection since we wanted to train a morph detector that performs well across different morphing techniques. In addition, after our network is fully trained, we zero out any grouped weight with a weight parameters norm smaller than 0.001. It was found that $\lambda = 0.003$ yields the top 20 most discriminative sub-bands, with the corresponding highest AUC of 99.31\%. Figure 3 depicts the selected sub-bands along with the corresponding numbers after training our network using universal dataset, and $\lambda= 0.003$.

Considering the aforementioned hyperparameter tuning process, we displayed the performance of our morph detector for some selected values of $\lambda$ using the validation portion of the universal dataset. Figure 4 shows the Detection Error Trade-off (DET) curves corresponding to different group sparsity regularization parameter $\lambda$. We can clearly see that the performance of our deep morph detector is at its best when $\lambda=0.003$.


\begin{table}[t]
\caption[Table caption text]{Performance of single morph detection: D-EER\%, BPCER@APCER=5\%, and BPCER@APCER=10\%.}      
\small
\begin{center}
\addtolength{\tabcolsep}{-0pt}
\begin{tabular}{lccccc}
\hline
Train&Test&Algorithm&D-EER&5\%&10\%\\ 
\hline
\multirow{28}{*}{\rotatebox[origin=c]{90}{Universal(VISAPP17+MorGAN+LMA)}}&\multirow{7}{*}{\rotatebox[origin=c]{90}{VISAPP17}}&BSIF+SVM~ \cite{kannala2012bsif}&                      35.00 &67.20&59.00\\
                                &&SIFT+SVM~ \cite{lowe1999object}&     27.00 &83.20&70.90\\
                                &&LBP+SVM~ \cite{ojala1994performance}&  37.67 &72.50 & 59.50\\
                                &&SURF+SVM~ \cite{bay2006surf}&  31.00 &79.40 & 70.10\\
                                
                                &&RGB+DNN~ \cite{szegedy2017inception}&  0.00 &0.00 & 0.00\\
                                &&48-sub-bands & 0.00 & 0.00 & 0.00\\
                                &&\textbf{Ours}&                                 {\bf0.00} &{\bf0.00}&{\bf0.00}\\\cline{2-6} 
                               
&\multirow{7}{*}{\rotatebox[origin=c]{90}{LMA}}&BSIF+SVM~ \cite{kannala2012bsif}&                      30.00 &70.42&57.60\\
                                &&SIFT+SVM~ \cite{lowe1999object}&     28.31 &67.70&50.00\\
                                &&LBP+SVM~ \cite{ojala1994performance}&  29.00 &61.50 & 51.20\\
                                &&SURF+SVM~ \cite{bay2006surf}&  33.40 &74.50 & 62.70\\
                                &&RGB+DNN~ \cite{szegedy2017inception}&  7.80 &13.00 & 6.10\\
                                &&48-sub-bands & 4.62 & 4.22 & 2.73\\
                                &&\textbf{Ours}&                                 {\bf4.44} &{\bf4.11}&{\bf2.21}\\\cline{2-6}

&\multirow{7}{*}{\rotatebox[origin=c]{90}{MorGAN}}&BSIF+SVM~\cite{kannala2012bsif}&                      28.80 &62.42&45.70\\
                                &&SIFT+SVM~ \cite{lowe1999object}&     47.60 &92.30&88.60\\
                                &&LBP+SVM~ \cite{ojala1994performance}&  31.20 &62.00 & 55.60\\
                                &&SURF+SVM~ \cite{bay2006surf}&  38.67 &76.00 & 70.00\\
                                
                                &&RGB+DNN~ \cite{szegedy2017inception}&  4.69 &4.70 & 2.74\\
                                &&48-sub-bands & 1.11 & 0.43 & 0.34\\
                                &&\textbf{Ours}&                                 {\bf1.53} &{\bf0.32}&{\bf0.30}\\\cline{2-6}

&\multirow{7}{*}{\rotatebox[origin=c]{90}{Universal}}&BSIF+SVM~ \cite{kannala2012bsif}&                      23.74 &51.42&38.67\\
                                &&SIFT+SVM~ \cite{lowe1999object}&     37.21 &87.45&76.71\\
                                &&LBP+SVM~ \cite{ojala1994performance}&  38.80 &91.36 & 83.40\\
                                &&SURF+SVM~ \cite{bay2006surf}&  36.00 &75.50 & 65.76\\
                                &&RGB+DNN~ \cite{szegedy2017inception}&  5.57 &6.08 & 3.00\\
                                &&48-sub-bands & 3.12 & 1.78 & 0.97\\
                                &&\textbf{Ours}&                                 {\bf2.78} &{\bf1.75}&{\bf1.21}\\\cline{2-6}

\end{tabular}
\end{center}

\label{table:results_cross}
\end{table}

\subsection{Grouped Weights Decay}
To show the functionality of the group sparsity regularization coefficient $\lambda$, we plot the norms of the grouped weights in the first convolutional layer of our DNN as a function of the group sparsity regularization coefficient $\lambda$ for some of the selected wavelet sub-bands (see Figure 5). Increasing the group sparsity penalty coefficient $\lambda$ will push the grouped weight norms toward zero as expected.


\begin{figure*}[t]
\begin{subfigure}{0.5\textwidth}
\begin{center}
\includegraphics[width=1\linewidth]{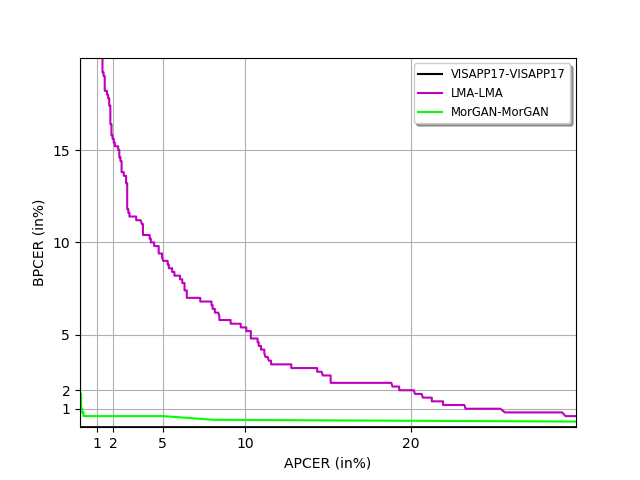}
\end{center}
\caption{}
\label{fig:DET1}
\end{subfigure}
\begin{subfigure}{0.5\textwidth}
\begin{center}
\includegraphics[width=1\linewidth]{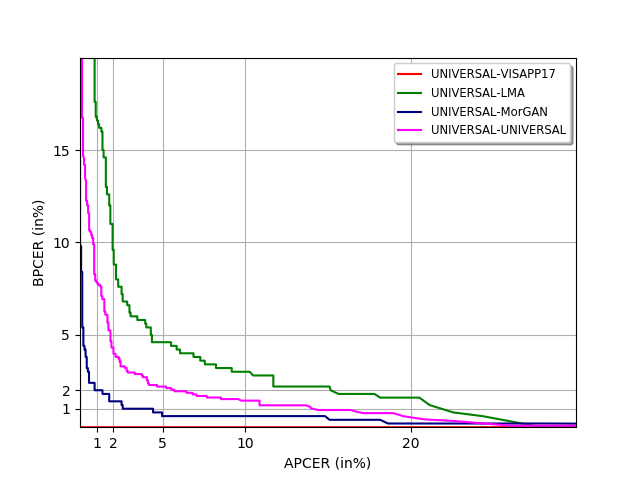}
\end{center}
\caption{}
\label{fig:DET2}
\end{subfigure}
\caption{ DET curves which display the performance of our morph detector when (a) trained and evaluated on individual datasets and (b) trained on the universal dataset.}
\label{fig:DET}
\end{figure*}

\subsection{Performance of the Deep Morph Detector}
 Once we found the top 20 discriminative wavelet sub-bands, we stacked the selected 20 wavelet sub-bands as data samples, and we retrained our modified Inception-ResNet-v1. It should be pointed out that the number of channels is reduced to 20. In the first scenario, we train the DNN using the 20-stacked wavelet sub-bands of each individual dataset, and the performance of the our morph detector is reported using the evaluation metrics introduced in Section 4.4. Table 1 delineates the performance of our deep morph detector when evaluated on our three datasets. Moreover, the corresponding DET curves are shown in Figure 6.a.
 
 In the second scenario, we trained our DNN detector using the training portion of the universal dataset, and the performance of the deep morph detector is reported using the test set of the universal dataset, as well as the test sets of individual datasets. Table 2 summarizes the evaluation of our morph detector when trained on the universal dataset, and the pertinent DET curves are plotted in Figure 6.b. 
 
 Please note that in Table 1 and Table 2, RGB+DNN represents our baseline when the Inception-ResNet-v1 is trained on the original RGB images, and 48-sub-band data indicates the data samples that consist of 48 wavelet sub-bands without utilizing the structured group sparsity. Our results mentioned in Table 1 and Table 2 prove the effectiveness of feature selection scheme for detecting morphed samples. In particular, the morphed samples in the VISAPP17 and MorGAN datasets are detected precisely compared with the LMA dataset. Please note that the performance of our RGB+DNN baseline, which is trained on the Inception-ResNet-v1, is on par with the performance of other DNNs. In particular, regarding the state-of-the-art results, in \cite{debiasi2019detection} the morph detection results on the LMA dataset is as follows: D-EER: 0.00, BPCER@APCER=10\%: 0.00 , BPCER@APCER=20\%: 0.00. Also, their results on the MorGAN dataset is as follows: D-EER: 34.00, BPCER@APCER=10\%: 67.00, BPCER@APCER=20\%: 78.00. Comparing the state-of-the-art results with ours reveals that our method outperforms on the MorGAN dataset considerably and perform close to this baseline on the LMA dataset.

\section{VISUALIZING THE FUNCTIONALITY OF THE DEEP MORPH DETECTOR}
In this section, we adopt a few visualization techniques to explain the underlying mechanism of our morph detector. First, we utilize the t-distributed Stochastic Neighbor Embedding (t-SNE) \cite{van2008visualizing} visualization technique to explain the classification performance improvement due to the imposed structured sparsity. Second, we utilize the Gradient-weighted Class Activation Mapping (Grad-CAM) \cite{selvaraju2017grad} to show the most attended spatial regions in the input images when our trained classifier labels an input image as a bona fide or morphed image.

\subsection{Visualizing the Functionality of Structured Group Sparsity}
To display the efficacy of our feature selection scheme, which is selecting the most discriminative sub-bands, the t-SNE visualization technique is employed as a representative medium which preserves the local structure of samples when visualizing high dimensional data samples in a low dimensional space. 
We randomly select 200 bona fide, and 200 morphed samples from the MorGAN dataset for the following two cases. In the first case, we extract the DNN embedding features for the original 48-sub-band data samples for the 200 bona fide and 200 morphed sample. Please note that the employed DNN for feature extraction was already trained on the 48-sub-band data. As the second case, we find the deep features using a DNN trained on the 20-sub-band data for the same 200 bona fide and 200 morphed samples. The point here is that we use the same data samples for both cases, but with different number of wavelet sub-bands. These two subsets of data points are plotted using t-SNE as shown in Figure 7, and we see that 20-sub-band data, in the right column, are more separable compared to the 48-sub-band data in the left column, which substantiates the effectiveness of our feature selection algorithm.

\subsection{Grad-CAM Visualization}
Understanding the key spatial areas in an input image, in terms of detection or classification, has been a long-standing topic of interest in the vision community. It is worth mentioning that our DNN is a non-attention-based architecture, and we do not use any attention mechanism in our DNN. In this section we adopt another useful visualization technique to observe which regions in the input images are paid more attention to from the DNN perspective in time of inference. In other words, we want to see which pixels are considered discriminative given our morph classification task. To explain the functionality of our trained DNN, we employ the Grad-CAM, which represents the gradient-weighted class activation maps.
In this visualization method, the gradient of a class-specific logit is obtained with respect to all spatial locations in a given feature map of the last convolutional layer in the DNN under scrutiny. The calculated gradients are averaged-pooled globally for each feature map, and these coefficients are used for a weighted average of the feature maps along with a final ReLU activation function to produce the class specific Grad-CAM. In accordance with the notation used in \cite{selvaraju2017grad}, the importance of the weights incorporating the pixels in the feature map $k$ of the last convolutional layer is as follows:
\begin{equation}
\alpha _k ^ {class} = \frac{1}{Z}\sum_i \sum_j \frac{\partial y^{class}}{\partial A_{ij}^{k}},  
\end{equation}
where $Z$ represents the total number of spatial locations $ij$ in the feature map $k$, $y^{class}$ delineates the score or logit for the class $class$, and $A$ denotes the activation or feature map. Consequently, the Grad-CAM produced for class $class$ with respect to the final convolutional layer in the DNN is as follows:
\begin{equation}
L _{Grad-CAM} ^ {class} = ReLU ( \sum_k  \alpha _k ^ {class}  A^{k} ).
\end{equation}
To produce the class-specific Grad-CAM for a bona fide image and its corresponding morphed image, we choose the last convolutional layer in our modified Inception-ResNet-v1, which has 1792 feature maps with spatial size of $3 \times 3$.  Based on Figure 8, the right images, that are morphed faces have substantially more attended regions compared to the left images, which are the bona fide samples.



\begin{figure}[t]
    
    \begin{center}
    \includegraphics[width=0.99\linewidth]{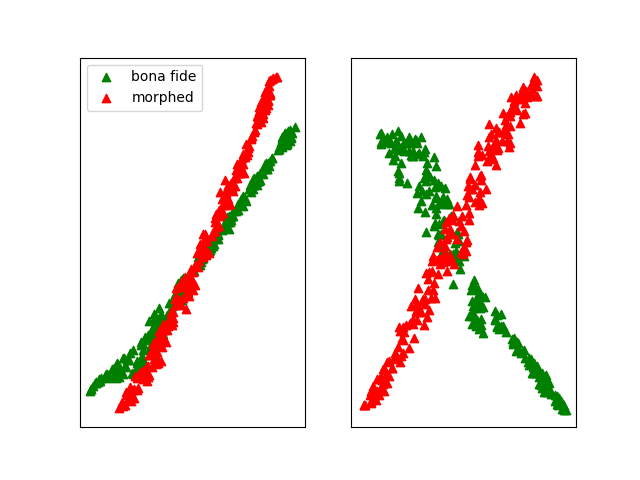} 
    \end{center}
    \caption{T-SNE visualization. The left figure depicts the 48-sub-band data, and the right figure shows the 20-sub-band data from the MorGAN dataset.}
\end{figure}

\begin{figure}[t]
    \begin{center}
    \includegraphics[width=0.55\linewidth]{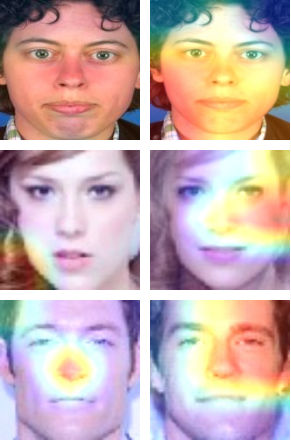} 
    \end{center}
    \caption{Grad-CAM visualizations for bona fides (left) and the corresponding morphed images (right). The first, second, and third rows represent samples from the VISAPP17, LMA, and MorGAN datasets, respectively. }
\end{figure}

\section{CONCLUSION}

In this paper, we employed structured group sparsity to force a DNN into finding the most discriminative subset of wavelet sub-bands, that are the wavelet sub-bands. To isolate the discriminating artifacts in the spatial-frequency feature domain, we adapted our framework into the wavelet domain. As far as learning the parameters of our DNN is concerned, the cost function of the DNN is constrained to meet the group Lasso condition, which is imposed on the grouped weights of the first convolutional layer. Our adjusted cost function results in finding the top 20 discriminative wavelet sub-bands, further enabling accurate morph detection with respect to our datasets. The D-EER, APCER5, and APCER10 rates obtained using our trained network with the optimal number of sub-bands substantiate the effectiveness of our framework. In particular, the morphed samples in the VISAPP17, and MorGAN datasets are detected accurately compared to the LMA dataset. In addition, to make the  effectiveness of our morph detector transparent, we utilized two visualization techniques to explain the functionality of the proposed single image morph detector. 

{\small
\bibliographystyle{ieee_fullname}
\bibliography{egbib}
}

\end{document}